\documentclass[lettersize,journal]{IEEEtran}
\usepackage{amsmath,amsfonts}
\usepackage{algorithmic}
\usepackage{algorithm}
\usepackage{array}
\usepackage[caption=false,font=normalsize,labelfont=sf,textfont=sf]{subfig}
\usepackage{textcomp}
\usepackage{stfloats}
\usepackage{url}
\usepackage{verbatim}
\usepackage{graphicx}
\usepackage{cite}
\usepackage{booktabs}
\usepackage{multirow}
\usepackage[table]{xcolor}
\usepackage[hidelinks]{hyperref}

\begin{document}

\title{From Sharp Eyes to Expert Mind: Internalizing Expert Knowledge \\ 
in MLLMs for Tampered Text Detection}

\author{
    Kaiqing Lin$^{\dagger}$,
    Songze Li$^{\dagger}$,
    Shen Chen$^{\dagger}$,
    Yunfei Guo, 
    Xiaoye Qiu, 
    Haodong Li,~\IEEEmembership{Member,~IEEE,} 
    Taiping Yao, 
    Bo Wang, 
    Youchang Xiao, 
    Bin Li$^{*}$,~\IEEEmembership{Senior Member,~IEEE,}
    and~Shouhong~Ding
    \thanks{
        $^{\dagger}$ Equal Contribution. $^{*}$ Corresponding author: Bin Li (email: libin@szu.edu.cn).
    }
    \thanks{
        Kaiqing Lin, Songze Li, Xiaoye Qiu, Haodong Li, and Bin Li are with 
        Guangdong Provincial Key Laboratory of Intelligent Information Processing, Shenzhen Key Laboratory of Media Security, and SZU-AFS Joint Innovation Center for AI Technology, Shenzhen University, Shenzhen 518060, China. (e-mails: linkaiqing2021@email.szu.edu.cn; 2310433002@email.szu.edu.cn; libin@szu.edu.cn).
        Shen Chen, Yunfei Guo, Taiping Yao, Bo Wang, Youchang Xiao, and Shouhong Ding are with Tencent Youtu Lab, Shanghai, China.
    }
}

\maketitle

\begin{abstract}
Tampered Text Detection (TTD) is essential for safeguarding document authenticity in security-critical workflows. Existing expert models are effective at capturing subtle manipulation traces but often generalize poorly across diverse document domains, while Multimodal Large Language Models (MLLMs) offer stronger semantic understanding and transferability yet remain insensitive to fine-grained forensic artifacts. This complementarity motivates us to investigate how expert forensic perception can be internalized into an MLLM rather than merely accessed through an external module.
We identify a fundamental \textbf{Double Mismatch} that hinders this goal: a \textit{Spatial Precision Mismatch} between coarse visual tokens and tiny tampered regions, and a \textit{Perceptual Granularity Mismatch} between semantics-oriented pre-training and low-level forensic perception. To address these challenges, we propose \textbf{Expert Knowledge Internalization (EKI)}, a progressive two-stage framework that transfers forensic expertise into the MLLM itself. In Stage 1, Text-Focused and Image-Focused strategies establish precise spatial focus on small text regions. In Stage 2, the proposed Forensic-General Representation Alignment (FGRA) loss aligns shallow LLM representations with those of a pre-trained forensic expert, enabling the model to acquire fine-grained artifact perception before such cues are diluted by deeper semantic abstraction. Extensive experiments on multiple in-domain and cross-domain benchmarks demonstrate that EKI achieves state-of-the-art performance and stronger generalization than existing expert-model-based and MLLM-based methods. Moreover, the expert is required only during training, allowing the resulting MLLM to maintain inference efficiency nearly identical to the vanilla model without relying on any external expert at inference.
\end{abstract}


\section{Introduction}

\IEEEPARstart{D}{ocument} images are widely used in financial, governmental, and legal workflows, where textual content supports critical decisions. However, increasingly accessible manipulation techniques enable attackers to alter critical textual information while preserving a visually plausible appearance, posing serious threats to document authenticity and information integrity. Detection of such manipulation is therefore essential for document forensics. Tampered Text Detection (TTD) aims to identify manipulated content and accurately localize corresponding regions in document images.

Existing TTD methods predominantly rely on visual backbones to build expert models, i.e., specialized forensic models, for detecting subtle traces left by manipulation operations. Representative architectures span CNN-based and hybrid spatial-frequency networks \cite{qu2023towards, chen2024enhancing, wong2025adcd}, Vision Transformer (ViT)-based models \cite{li2025ditl2}, and more recent Mamba-based architectures \cite{duan2025realdtt}. These methods typically formulate TTD as binary classification or pixel-level segmentation and train their models on task-specific collections of authentic and tampered document images. Such training enables expert models to learn effective low-level forensic cues and achieve strong in-domain performance. However, existing training datasets often cover only a limited range of document distributions. Consequently, expert models may overfit to domain-specific visual characteristics, limiting their generalization to unseen domains \cite{li2025ditl2}.

Recently, Multimodal Large Language Models (MLLMs) have emerged as a promising foundation for generalizable document forensics. Benefiting from large-scale vision-language pre-training, MLLMs exhibit strong transferability across diverse visual domains and can better adapt to variations in document content, layout, and appearance \cite{bai2025qwen2, ye-etal-2023-ureader, fujitake-2024-layoutllm}. Beyond cross-domain generalization, MLLMs also provide visual content understanding and multimodal reasoning capabilities. Recent studies have begun to explore these advantages for tampered text analysis. TextSleuth \cite{qu2024textsleuth} introduces natural-language explanations for tampering evidence and improves fine-grained analysis through a two-stage paradigm that focuses on suspected regions. TVSIP \cite{xu2025pixels} further integrates visual localization with semantic interpretation, enabling joint tampering detection and explainable forensic analysis within an MLLM-driven framework. These studies demonstrate the potential of MLLMs to extend conventional forensic localization toward more comprehensive visual-semantic analysis. However, existing MLLM-based methods still lag behind expert models in detecting and localizing subtle, fine-grained tampering.

\begin{figure}[t]
\centering
\includegraphics[width=\columnwidth]{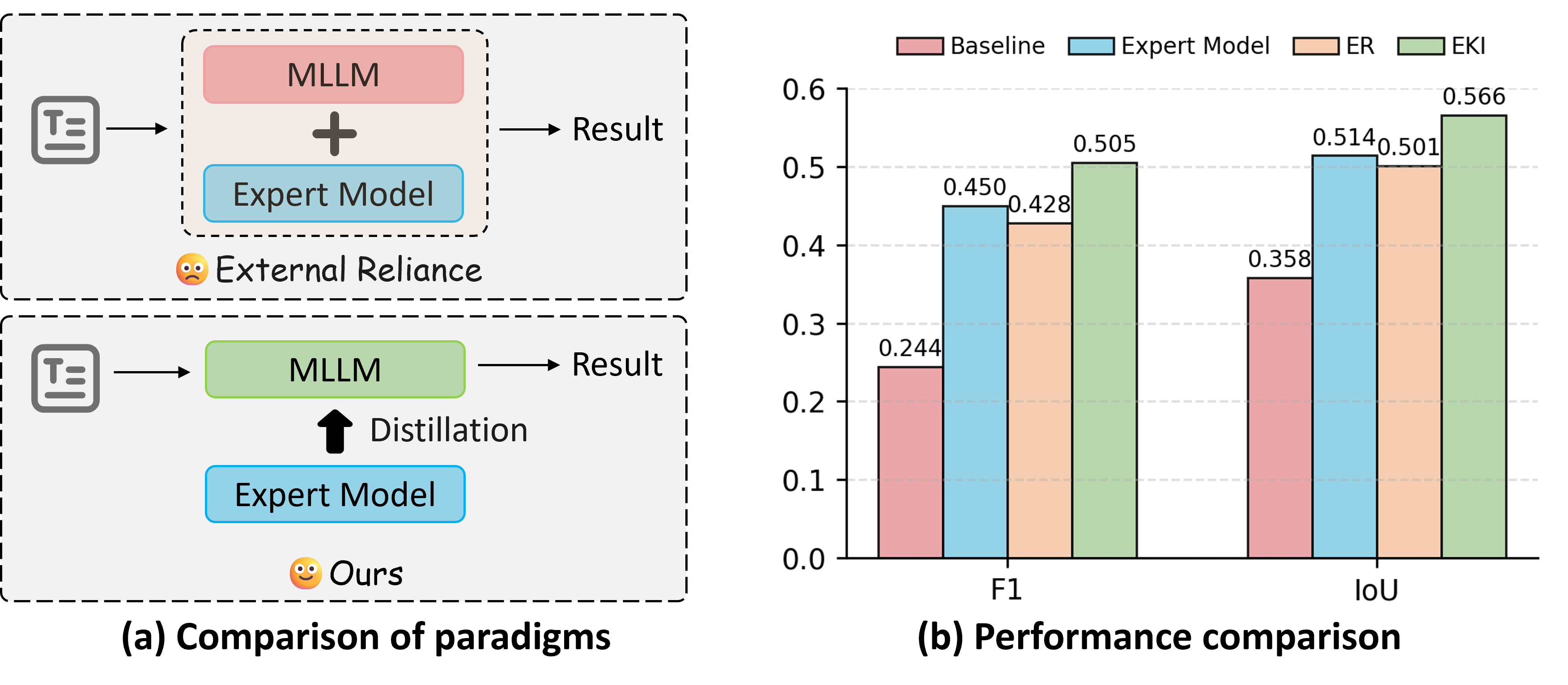}
\caption{(a) Unlike external-reliance methods, ours internalizes forensic knowledge for intrinsic authenticity detection without extra modules. (b) Performance of Baseline (standard fine-tuned MLLM), Expert Model \cite{li2025ditl2}, ER (External Reliance), and our EKI (Expert Knowledge Internalization). Notably, EKI significantly outperforms other paradigms, demonstrating the superiority of intrinsic forensic capabilities.}
\label{fig:introduction}
\end{figure}

Expert models and MLLMs thus exhibit strikingly complementary strengths: the former perceive the subtle manipulation traces that the latter miss, while the latter provide the document understanding and cross-domain generalization that the former lack. This contrast raises the central question of this paper: \textbf{how can an MLLM be equipped with the expert's forensic perception, such that a single model perceives like a specialist while understanding and generalizing like a large foundation model?}
A natural way to exploit this complementarity is to integrate an expert model into the MLLM pipeline.
Existing approaches commonly follow such an external-reliance paradigm \cite{zhang2024common, chaubey2025face, zhou2025aigi}, keeping the expert as a separate module and coupling it with the MLLM via late fusion or prediction concatenation (Fig. \ref{fig:introduction}(a), top). 
However, our comparative study reveals a notable limitation of this paradigm: as shown in Fig. \ref{fig:introduction}(b), the externally coupled system fails to surpass the standalone expert, despite requiring a more complex and costly inference pipeline. 
In other words, external access to expert predictions does not translate into intrinsic forensic perception within the MLLM.

\begin{figure}[t]
\centering
\includegraphics[width=\columnwidth]{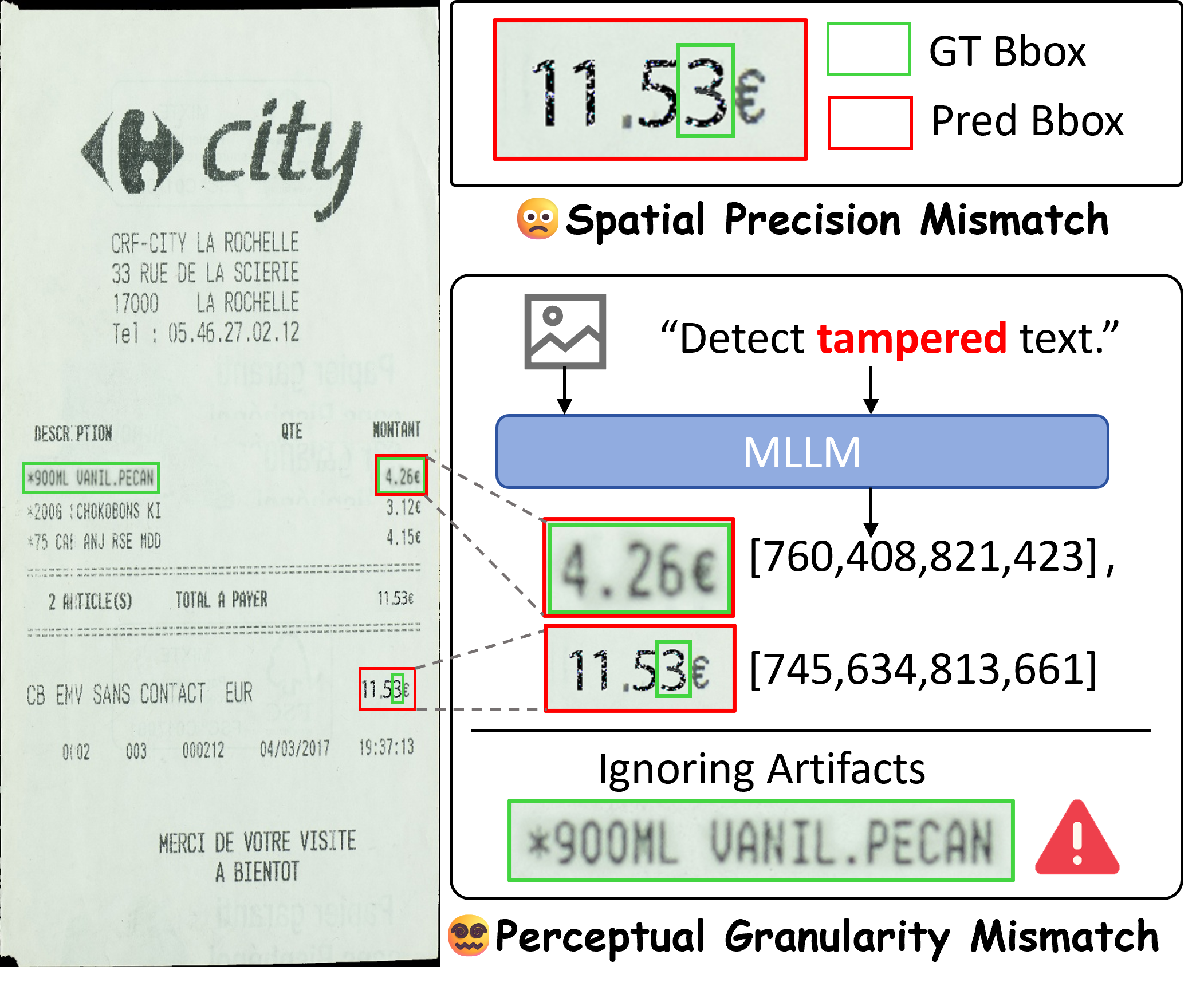}
\caption{General MLLMs suffer from the Double Mismatch: Spatial Precision Mismatch (difficulty in precisely localizing tiny regions) and Perceptual Granularity Mismatch (understanding semantics but remaining blind to fine-grained artifacts).}
\label{fig:double_mismatch}
\end{figure}

Why does the coupling fall short? Under external reliance, the expert’s predictions or features are directly inserted into the pipeline, while the MLLM has limited opportunity to learn the underlying forensic representations itself. As a result, the coupled system largely inherits the expert’s forensic capability and its limited cross-domain generalization, without fully integrating the MLLM’s semantic understanding and generalization strengths into forensic representation learning.
Nor can the MLLM acquire such perception through standard fine-tuning alone: the fine-tuned Baseline in Fig. \ref{fig:introduction}(b) lags far behind the expert. We attribute this learning difficulty to a fundamental \textbf{Double Mismatch} (Fig. \ref{fig:double_mismatch}). 
First, a \textbf{Spatial Precision Mismatch}: character-level tampered regions occupy only a tiny fraction of the visual token sequence, making the relevant evidence both spatially under-resolved and easily overwhelmed by irrelevant background during optimization. 
Second, a \textbf{Perceptual Granularity Mismatch}: MLLM representations are optimized for high-level vision-language semantics through objectives such as contrastive image-text alignment and autoregressive prediction \cite{radford2021learning, zhai2023sigmoid, bai2025qwen2}, while recent studies in image forensics show that such semantics-oriented representations are often insufficiently sensitive to low-level forensic signals \cite{lin2025seeingreasoningunifiedframework, lin2026deep}.
Hence, neither borrowed predictions nor standard fine-tuning suffices; what the MLLM needs is guidance: expert knowledge that steers its own representation learning.

To this end, we propose \textbf{Expert Knowledge Internalization (EKI)} (Fig. \ref{fig:introduction}(a), bottom): rather than feeding the expert's predictions to the MLLM in textual or visual form, or attaching the expert as an additional visual encoder to inject its features, EKI employs it as a teacher that guides the MLLM to learn forensic perception within its own representations, via a progressive two-stage framework targeting the two mismatches. Stage 1 (\textbf{Precise Spatial Focus}) tackles the spatial mismatch, teaching the model \textit{where to look}: Text-Focused and Image-Focused strategies force attention onto tiny text regions, so that subsequent learning concentrates on the true evidence. 
Stage 2 (\textbf{Forensic Artifact Perception}) tackles the perceptual mismatch, teaching the model \textit{how to discern}: the proposed Forensic-General Representation Alignment (FGRA) loss aligns the LLM's shallow-layer representations with those of a pre-trained forensic expert, where fine-grained forensic cues are better preserved.
Since the forensic capability is internalized into the MLLM itself, the expert is discarded after training, introducing no additional inference overhead. As shown in Fig. \ref{fig:introduction}(b), the resulting model outperforms both the external-reliance paradigm and the standalone expert, while exhibiting stronger cross-domain generalization.

Our contributions are summarized as follows:
\begin{itemize}
    \item We reveal the limitations of existing expert–MLLM integration paradigms and identify the \textbf{Double Mismatch} that prevents general MLLMs from acquiring fine-grained forensic perception, motivating a shift from external reliance to intrinsic expert knowledge internalization.

    \item We propose the \textbf{Expert Knowledge Internalization (EKI)} paradigm, a progressive two-stage framework that first establishes precise spatial focus and then transfers fine-grained forensic perception from a pre-trained expert into the MLLM.

    \item Extensive experiments demonstrate state-of-the-art performance and superior cross-domain generalization over existing expert-model-based and MLLM-based methods, while maintaining efficiency nearly identical to the vanilla MLLM and requiring no external expert at inference.
\end{itemize}

\section{Related Work}

\subsection{Document Text Tampering}
Document images in financial, governmental, and legal workflows carry security-sensitive text such as amounts, dates, and names, making them prime targets for manipulation. Following representative TTD datasets \cite{qu2023towards, yu2025toward}, document text tampering comprises five representative types (Table \ref{tab:dataset_stats}): \textbf{Copy-move} duplicates text within the same image, so the forged region shares font, illumination, and compression statistics with its surroundings; \textbf{Splicing} pastes text from a different image, introducing cross-image inconsistencies in noise or compression history; \textbf{Removal} erases text and inpaints the background, hiding information without adding new content; \textbf{Insertion} renders new text into blank regions; and \textbf{Replacement} erases the original text and inserts new content in place, which is the most prevalent and harmful type in practice. Detecting such tampering is challenging: manipulated regions are often as tiny as individual characters, their traces become visually imperceptible after post-processing such as blurring, noise addition, and recompression \cite{yu2025toward}, and forgery styles vary widely across document domains \cite{comp2020, comp2022}. Consequently, most existing detection methods resort to fine-grained, low-level forensic features for detection, as reviewed below.

\subsection{General Image Manipulation Localization}
Recent years have witnessed rapid advancements in general image manipulation localization \cite{liu2022pscc, ma2023iml, zeng2023toward, guillaro2023trufor, chen2024ean, liu2025image, su2025can, zhu2025mesoscopic, 11155891, 9393396, 10223250, 11570902}. To capture diverse tampering artifacts, various architectures have been explored. For instance, PSCC-Net \cite{liu2022pscc} introduces bidirectional pathways to effectively aggregate multi-scale features, while IML-ViT \cite{ma2023iml} focuses on high-resolution modeling to improve boundary awareness. TruFor \cite{guillaro2023trufor} employs a transformer-based framework to fuse high-level semantics from RGB images with low-level traces derived from learned noise-sensitive fingerprints. EAN \cite{chen2024ean} formulates the task as a boundary-aware segmentation problem, utilizing an edge interaction mechanism to refine both implicit and explicit contour features. To better isolate manipulation traces from normal image content, SparseViT \cite{su2025can} decouples semantic and non-semantic features, adaptively concentrating on the non-semantic cues that are crucial for forensics. Similarly, Mesorch \cite{zhu2025mesoscopic} captures local tampering traces and global inconsistencies by integrating distinct frequency components and hierarchical features through an adaptive weighting scheme. 
Despite their remarkable success, these methods are fundamentally tailored for natural scene images. Consequently, they often exhibit suboptimal performance when directly applied to document images due to the lack of domain-specific text priors and the inability to handle dense textual interference.

\subsection{Tampered Text Detection (TTD)}
\subsubsection{Expert Models}
Traditional TTD methods have evolved from handcrafted feature-based approaches to deep learning frameworks. To capture subtle low-level forensic traces, researchers have explored complementary forensic cues beyond raw RGB information. For instance, DTD \cite{qu2023towards} and FFDN \cite{chen2024enhancing} leverage frequency-domain analysis to mine spectral inconsistencies that are difficult to perceive directly in RGB images, while ADCD-Net \cite{wong2025adcd} combines adaptive DCT features with OCR-guided \cite{CRAFT,cui2025paddleocr} content disentanglement. 
BOIL \cite{10330057} further explores frequency-aware data augmentation to improve the robustness of forensic feature learning.
Beyond spectral cues, dual-branch architectures such as TIFDM \cite{dong2024robust} fuse RGB information with hand-crafted residuals (e.g., SRM \cite{zhou2018learning,srm}) to capture statistical forgery traces at the noise level. Rather than introducing auxiliary modalities, DCL-Net \cite{li2026document} incorporates contrastive learning to enhance feature discriminability. From a data and pre-training perspective, DITL$^2$ \cite{li2025ditl2} improves cross-domain generalization by pre-training on diverse images to learn domain-invariant representations transferable to document forensics. Despite these advances, expert models remain largely task-specific, with limited adaptability to diverse and unseen document distributions and limited capability to exploit the rich textual and structural information inherent in documents. These limitations motivate the exploration of MLLM-based approaches.

\subsubsection{MLLM-based Approaches}
Benefiting from large-scale vision-language pre-training, MLLMs exhibit strong transferability across diverse visual domains and provide visual-semantic understanding and multimodal reasoning capabilities, offering a promising foundation for document forensics \cite{bai2025qwen2, ye-etal-2023-ureader, fujitake-2024-layoutllm}. Motivated by this potential, recent studies have begun to explore MLLMs for tampered text analysis, progressively extending their role from forensic interpretation to tampering localization. TextSleuth \cite{qu2024textsleuth} employs an external expert model to first localize suspicious regions and then guides the MLLM to explain the corresponding tampering evidence, leaving the MLLM itself largely uninvolved in detection and localization. TVSIP \cite{xu2025pixels} goes a step further by introducing an MLLM-based semantic branch for localization and fusing its predictions with those of an external forensic expert.
However, TVSIP performs only prediction-level fusion between independently modeled branches, without feature-level interaction or joint representation learning. 
Consequently, current methods do not internalize the expert's fine-grained forensic perception, and the external expert remains necessary during inference. 
More broadly, existing MLLMs remain insensitive to subtle forensic artifacts \cite{lin2025seeingreasoningunifiedframework, lin2026deep}, which expert models are better specialized to capture. This complementarity motivates us to internalize the expert's forensic perception into the MLLM.

\subsection{Integrating MLLMs with Expert Models}

\subsubsection{External Reliance Strategies}
Vanilla MLLMs remain ineffective at detecting subtle image forgeries due to their limited sensitivity to fine-grained forensic artifacts \cite{11124461}.
To mitigate the perceptual limitations of MLLMs, many studies integrate external experts. FFAA \cite{huang2024ffaa} employs a decision system to select responses based on matching scores, while $\mathcal{X}^2$-DFD \cite{chen2024x2} calls external detectors when encountering ambiguous features. AIGI-Holmes \cite{zhou2025aigi} incorporates an NPR-based \cite{tan2024rethinking} visual expert to extract artifacts. While improving performance, these approaches rely heavily on external modules, reducing the MLLM to a passive reasoning interface.

\subsubsection{Knowledge Internalization Strategies}
A more efficient direction empowers MLLMs intrinsically via distillation. Recent works like SF \cite{li2025spatial} and VIRAL \cite{yoon2025visual} demonstrate that aligning the MLLM's feature space with a teacher expert injects domain knowledge without architectural changes. Inspired by this, we propose internalizing forensic knowledge into MLLMs. Unlike general visual alignment, our method specifically targets the fine-grained artifact features essential for TTD.

\section{Preliminaries}

\subsection{Task Reformulation}
\label{subsec:reformulation}
Traditionally, tampered text detection has been formulated as a dense semantic segmentation task, where expert models aim to generate a binary mask classifying each pixel as authentic or manipulated. However, in practical forensic scenarios, the primary goal is often to identify which specific object or text region has been manipulated, rather than delineating its exact pixel boundaries. Motivated by this, and recognizing the remarkable proficiency of MLLMs in visual grounding, we reframe the objective from pixel-level segmentation to coordinate-based \textbf{bounding box (bbox) regression}. Specifically, the ground-truth bboxes are derived directly from the pixel-level masks provided by existing datasets: for each tampered region, we take the minimum enclosing rectangle defined by the leftmost, topmost, rightmost, and bottommost points of its mask, represented as $[x_1, y_1, x_2, y_2]$, requiring no additional annotation. Compared with dense masks, bounding box coordinates can be naturally serialized into a short sequence of discrete tokens, making the output format inherently compatible with the autoregressive decoding paradigm of language models.

\begin{figure*}[t]
\centering
\includegraphics[width=\textwidth]{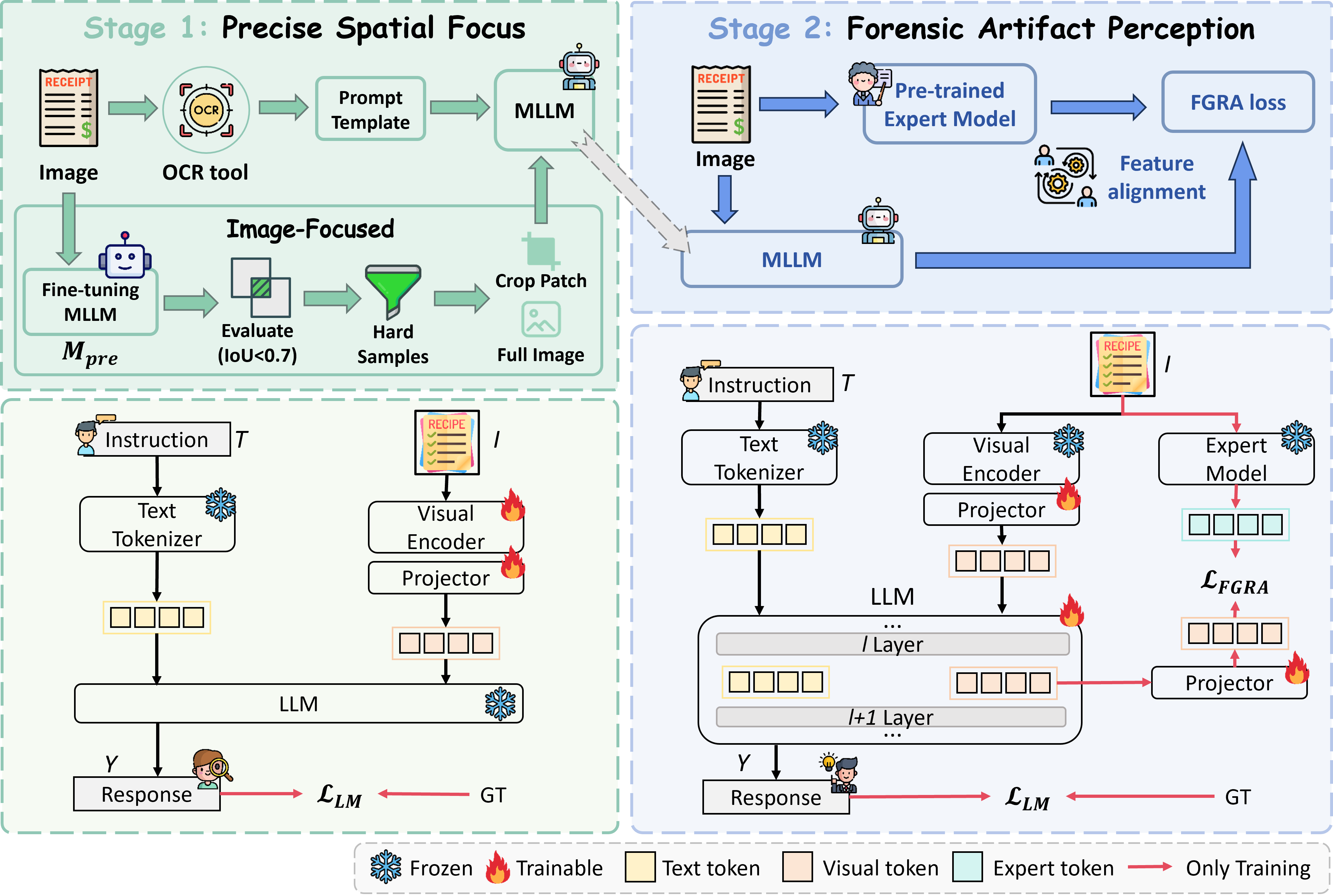}
\caption{Overview of our proposed \textbf{Expert Knowledge Internalization (EKI)} framework. \textbf{Stage 1: Precise Spatial Focus.} We enhance spatial precision using Text-Focused and Image-Focused strategies. \textbf{Stage 2: Forensic Artifact Perception.} The proposed FGRA Loss distills forensic knowledge from a pre-trained expert into the LLM's shallow layers, enabling intrinsic artifact perception. Note that the expert is discarded during inference.}
\label{fig:overview}
\end{figure*}

\subsection{Problem Definition}
\label{subsec:problem_def}
Formally, given a potentially tampered document image $\mathbf{I} \in \mathbb{R}^{H \times W \times 3}$ and a user text instruction $\mathbf{T}$ (e.g., ``\textit{Detect and localize the tampered text in this image}''), our goal is to generate a text response $\mathbf{Y}$. Unlike traditional masks, $\mathbf{Y}$ contains the sequence of bounding box coordinates $[x_1, y_1, x_2, y_2]$ corresponding to the manipulated regions, along with authenticity judgments.

\section{Methodology}

\subsection{Overview}
General MLLMs lack the fine-grained forensic perception required for TTD, yet neither externally coupling an expert model nor standard fine-tuning, which is hindered by the Double Mismatch, can effectively equip them with such capability. To address this problem, we propose Expert Knowledge Internalization (EKI), which employs a pre-trained forensic expert as a teacher to steer the MLLM's own representation learning. Specifically, EKI adopts a progressive two-stage training framework, as illustrated in Fig. \ref{fig:overview}, dismantling the two mismatches in turn: the model first learns \textit{where to look} (Stage 1) and then \textit{how to discern} (Stage 2). The design rationale is twofold:

\textbf{Stage 1: Precise Spatial Focus.} General MLLMs struggle to resolve tiny tampered regions due to coarse patch processing. Therefore, we dedicate this stage to forcing the model to localize precisely via explicit Text-Focused and Image-Focused strategies.

\textbf{Stage 2: Forensic Artifact Perception.} To overcome the insensitivity of the MLLM's semantics-oriented representations to subtle forensic artifacts, we introduce the Forensic-General Representation Alignment (FGRA) loss. This distills forensic capabilities from a pre-trained expert directly into the MLLM.
Crucially, the expert serves solely as a teacher during training and it is discarded at inference, ensuring the MLLM operates without additional overhead.

\subsection{Stage 1: Precise Spatial Focus}
The primary obstacle in the first stage is the \textit{Spatial Precision Mismatch}. 
To equip the model with fine-grained spatial attention capability, we choose to fine-tune the vision encoder and the projector in this stage, while keeping the LLM frozen.
Since document tampering often occurs at the character level, the tampered regions occupy a negligible portion of the image. Consequently, under coarse-grained visual tokenization, the overwhelming background semantics create significant noise interference, making direct regression difficult for standard MLLMs.
To address this, we employ a two-pronged strategy operating at both the \textit{data processing level} (via Text-Focused) and the \textit{training paradigm level} (via Image-Focused) to enhance the model's grounding capability.

\subsubsection{Text-Focused} 
To address the ambiguity of visual localization in complex documents, we construct a specialized instruction dataset (see Fig. \ref{fig:spatial_precision}(a)).
By using OCR-detected regions as coarse spatial anchors in the prompt and supervising the model to predict the precise bounding boxes $[x_1, y_1, x_2, y_2]$ aligned with ground-truth annotations, we force the MLLM to attend to the specific spatial tokens associated with text regions, rather than relying on general semantic contexts. This enables the model to perform a coarse-to-fine attention refinement, establishing the spatial foundation required for the subsequent forensic analysis. 

The data construction process is as follows:
\begin{itemize}
    \item \textbf{OCR Pre-processing.} We utilize an off-the-shelf OCR tool \cite{cui2025paddleocr} to extract semantic and spatial information from the document. The tool processes the image and outputs a set of detected text instances. For each instance, it provides three key components: the bounding box coordinates $\mathbf{b}_{ocr}$, the recognized text content $\mathbf{t}_{content}$, and a confidence score $\mathbf{s}_{conf}$.
    \item \textbf{High-Confidence Filtering.} The confidence score $\mathbf{s}_{conf}$, generated by the OCR tool, represents the model's predicted probability that the detected region contains valid text and that the recognition result is correct. To ensure the reliability of textual focus and suppress OCR noise, we empirically set a strict confidence threshold of $\mathbf{s}_{conf} > 0.9$ based on preliminary analysis, and retain only high-confidence OCR instances for instruction construction. We then randomly sample from this filtered pool to generate instructions.
    \item \textbf{Instruction Generation.} We construct the input instruction $\mathbf{T}$ by binding the content with its coarse location: 
    ``\textit{Is the text `$\mathbf{t}_{content}$' located at $\mathbf{b}_{ocr}$ real or tampered?}''
    \item \textbf{Target Formulation.} The ground-truth response $\mathbf{Y}$ depends on the intersection between the queried OCR region and actual tampering annotations:
    \begin{itemize}
        \item If real: ``\textit{Real.}''
        \item If tampered: ``\textit{Tampered. $\mathbf{b}_{gt}$.}'' 
    \end{itemize}
    Crucially, $\mathbf{b}_{gt}$ denotes the precise coordinates $[x_1, y_1, x_2, y_2]$ of the actual manipulated area (e.g., a specific modified digit) within the queried region, distinct from the coarse input $\mathbf{b}_{ocr}$. 
\end{itemize}

\subsubsection{Image-Focused}
\begin{figure}[t]
\centering
\includegraphics[width=\columnwidth]{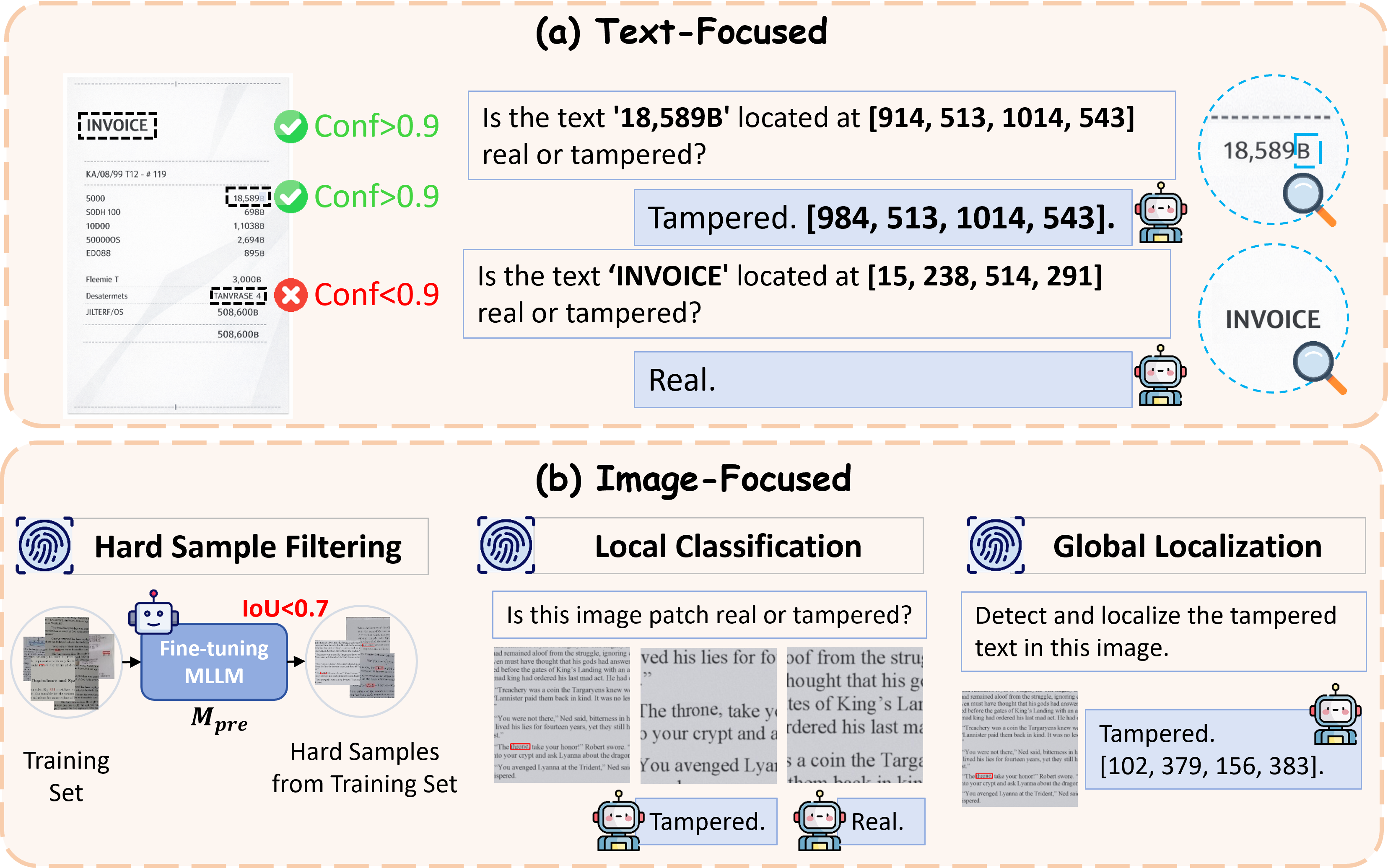}
\caption{Details of Precise Spatial Focus. We construct text-focused instructions using OCR results and employ a task decomposition strategy (Local Classification + Global Localization) to mine hard samples.}
\label{fig:spatial_precision}
\end{figure}

General MLLMs typically struggle to resolve fine-grained details in dense documents, making direct localization of tiny tampered regions (e.g., $<1\%$ of the image area) a formidable challenge. Nevertheless, existing training paradigms directly supervise the model with the full localization objective, overlooking this inherent task difficulty; as a result, hard samples contribute little effective learning signal. To mitigate this \textit{Spatial Precision Mismatch}, we propose a divide-and-conquer task decomposition strategy (as illustrated in Fig. \ref{fig:spatial_precision}(b)): we augment the hard samples with an easier classification sub-task on magnified local patches, thereby lowering the learning difficulty.
The core intuition is to simplify the learning problem: the model learns to distinguish tampering in isolated, magnified views, and this discriminative capability in turn facilitates localizing the same targets within the cluttered global context.
To achieve this, we design a three-step pipeline involving hard sample filtering and task decomposition:
\begin{itemize}
    \item \textbf{Hard Sample Filtering.} Rather than relying on static heuristics, we define ``hard samples'' in a data-driven manner based on the learning difficulty. Specifically, we first conduct a preliminary training round on the entire dataset using the standard MLLM baseline (fine-tuned for the global localization task) to obtain a temporary mining model, denoted as $\mathcal{M}_{pre}$. We then evaluate this model $\mathcal{M}_{pre}$ on the training set. Samples where the Intersection-over-Union (IoU) between the predicted and ground-truth boxes remains below a threshold of $0.7$ are identified as hard samples. These samples typically feature extremely tiny tampering regions or highly subtle manipulation traces that are easily overwhelmed by background semantic noise, making them persistent failure cases for standard training paradigms. These are then flagged for our specific decomposition training.
    \item \textbf{Local Classification (Sub-task 1).} For these identified hard samples, we dynamically crop both the tampered regions and random authentic regions to generate localized patches. Crucially, these patches are resized to the same resolution as the original full images (e.g., $512 \times 512$) before being fed into the MLLM. With the simplified instruction ``\textit{Is this image patch real or tampered?}'', the model can focus on discerning manipulation traces in these zoomed-in views without the interference of complex global backgrounds.
    \item \textbf{Global Localization (Sub-task 2).} Concurrently, we retain the original full-image localization task. By jointly training on these augmented local patches (Sub-task 1) and the original full images (Sub-task 2), the discriminative features learned from the magnified local views significantly enhance the model's ability to localize these difficult regions in the global view.
\end{itemize}

This decomposition strategy allows the two sub-tasks to reinforce each other: the easier local classification task provides a strong and direct learning signal for discerning manipulation traces, which in turn benefits the harder task of localizing them in cluttered full-page views.

\subsubsection{Training Objectives (Stage 1)}
To unify the learning of both sub-tasks, we employ the standard auto-regressive language modeling objective. Given the input instruction $\mathbf{T}$ and image $\mathbf{I}$, the model maximizes the likelihood of the target response sequence $\mathbf{Y} = \{y_1, \dots, y_L\}$. 
The standard text generation loss function $\mathcal{L}_{LM}$ is defined as:
\begin{equation}
    \mathcal{L}_{LM} = -\sum_{t=1}^{L} \log P(y_t | \mathbf{I}, \mathbf{T}, y_{<t})
\end{equation}
where $L$ is the length of the target sequence $\mathbf{Y}$, and $y_{<t}$ represents the preceding tokens. 

Our training strategy involves jointly optimizing the model on both hard-mined local patches $\mathcal{D}_{local}$ and global localization samples $\mathcal{D}_{global}$. The total objective for Stage 1 is formulated as:
\begin{equation}
    \mathcal{L}_{Stage1} = \mathcal{L}_{LM}^{local} + \mathcal{L}_{LM}^{global}
\end{equation}
where $\mathcal{L}_{LM}^{local}$ and $\mathcal{L}_{LM}^{global}$ represent the standard language modeling loss computed on the local and global datasets, respectively.
Crucially, the format of the target $\mathbf{Y}$ adapts to the sub-task:

For \textbf{Global Localization ($\mathcal{D}_{global}$):} $\mathbf{Y}$ contains both the judgment and coordinates (e.g., ``\textit{Tampered. $[x_1, y_1, x_2, y_2]$}'').

For \textbf{Local Classification ($\mathcal{D}_{local}$):} $\mathbf{Y}$ contains only the binary label (e.g., ``\textit{Tampered}'' or ``\textit{Real}'').

By minimizing this unified $\mathcal{L}_{Stage1}$, the Vision Encoder and Projector are updated to capture the fine-grained spatial features required for precise spatial focus.

\subsection{Stage 2: Forensic Artifact Perception}
With precise spatial focus established, the model has effectively learned to isolate the ``suspected'' regions from the overwhelming background noise, zooming in from the full image to character-level precision. However, mere localization is insufficient; the remaining challenge is the \textit{Perceptual Granularity Mismatch}—confirming whether these localized regions contain genuine tampering traces.  
In this stage, we freeze the Vision Encoder to preserve the spatial features learned in Stage 1, and instead fine-tune the LLM and the projector. This strategy facilitates the semantic alignment of forensic features within the language model's latent space. 

\subsubsection{Pre-trained Forensic Expert}
The choice of the expert model is pivotal for effective knowledge distillation. Inspired by recent advances in document forensics \cite{li2025ditl2}, we employ DINOv2 \cite{oquab2023dinov2} as our external forensic expert, as it is adept at capturing subtle low-level tampering traces, complementing the perceptual capability that general MLLMs lack.
To adapt this general-purpose backbone to the specific nuances of document forgery, we first fine-tune DINOv2 on our forensic training dataset. Once trained, we freeze its parameters throughout the subsequent internalization stage. 
Given an input image $\mathbf{I}$, this pre-trained expert $\mathcal{E}$ extracts visual feature representations, denoted as $\mathbf{F}_{exp} \in \mathbb{R}^{M \times D_{exp}}$, where $M$ represents the number of patches and $D_{exp}$ is the feature dimension. These features encode rich, generalized forensic representations, serving as a stable and high-quality supervision signal for the MLLM.

\subsubsection{Forensic-General Representation Alignment (FGRA)}
General MLLMs rely solely on the text generation loss $\mathcal{L}_{LM}$ to update visual representations. However, this textual supervision is indirect and sparse, failing to provide the explicit, dense feedback required to capture subtle tampering traces. This deficiency is a primary cause of the model's poor performance in fine-grained forensics. What the MLLM needs, therefore, is expert guidance that directly steers its own representation learning.
To provide such guidance, we introduce the FGRA loss, which provides direct supervision by distilling forensic knowledge into the LLM's representations.
Our rationale for targeting shallow layers lies in the hierarchical information processing of deep networks: while deep layers in LLMs evolve into highly abstract, semantically dominant representations, shallow layers act as the initial processing stage, retaining more structural and textural details (e.g., local patterns, high-frequency cues). These details, while less semantically meaningful, are strongly correlated with forensic artifacts. 
Therefore, the core idea is to align the shallow tokens of the LLM with the robust feature embeddings from the pre-trained DINOv2 expert, explicitly forcing these early layers to capture manipulation traces before they are diluted by subsequent semantic abstraction.
Specifically, to align with the expert features $\mathbf{F}_{exp}$, we extract the hidden states $\mathbf{H}^{(l)} \in \mathbb{R}^{N \times D_{llm}}$ from the $l$-th layer of the LLM, which correspond to the visual input tokens. Since the LLM and the expert operate in different semantic and dimensional spaces ($D_{llm} \neq D_{exp}$), we introduce a learnable projection head $\phi(\cdot)$ to map $\mathbf{H}^{(l)}$ into the expert's feature space.
For the spatial correspondence, the inputs of the two branches are unified in the preprocessing pipeline such that the MLLM's visual-token grid and the expert's patch grid share the same spatial sequence length (i.e., $N = M = K$). The alignment is thus a deterministic one-to-one, patch-wise pairing, where the $i$-th projected visual token is matched with the $i$-th expert patch feature.
The FGRA loss is formulated as the negative cosine similarity between the projected LLM tokens and the expert features:
\begin{equation}
    \mathcal{L}_{FGRA} = - \frac{1}{K} \sum_{i=1}^{K} \text{cos}(\phi(\mathbf{h}^{(l)}_i), \mathbf{f}_{exp, i})
\end{equation}
where $\mathbf{h}^{(l)}_i$ and $\mathbf{f}_{exp, i}$ denote the $i$-th token-feature pair, and $K$ is the shared spatial sequence length defined above.
By minimizing $\mathcal{L}_{FGRA}$, we force the aligned LLM layer to encode artifacts discernible by DINOv2, effectively ``internalizing'' the expert's eye for forgery.

\subsubsection{Training Objectives (Stage 2)}
The final training objective in the second stage combines the text generation loss with our proposed alignment loss. The total loss $\mathcal{L}_{Stage2}$ is formulated as:
\begin{equation}
    \mathcal{L}_{Stage2} = \mathcal{L}_{LM} + \mathcal{L}_{FGRA}
\end{equation}
where $\mathcal{L}_{LM}$ ensures the preservation of instruction-following, while $\mathcal{L}_{FGRA}$ injects the forensic discriminative power into the LLM's latent space.

\section{Experiments}

\subsection{Experimental Setup}

\subsubsection{Datasets}
To comprehensively evaluate the generalization capability of our method, we conduct experiments under both in-domain and cross-domain settings. 
As summarized in Table \ref{tab:dataset_stats}, we employ two distinct datasets for training: DocTamper-TrainingSet \cite{qu2023towards} and FSTS-T \cite{yu2025toward}. This dual-source training strategy allows us to verify the adaptability of our method to different data distributions. 
We utilize a diverse array of seven datasets for evaluation. Specifically, DocTamper-TestingSet serves as the in-domain test set for models trained on DocTamper-TrainingSet, while FSTS-S serves as the in-domain test set for FSTS-T. The remaining five datasets—DocTamper-FCD, DocTamper-SCD, FSTS-1.5K, SACP \cite{comp2020}, and RIFLC \cite{comp2022}—are used as cross-domain benchmarks to rigorously assess out-of-distribution generalization performance. 

\begin{table}[htbp]
\centering
\caption{Summary of datasets used in our experiments. Com, Spl, Rem, Ins, and Rep denote Copy-move, Splicing, Removal, Insertion, and Replacement, respectively.}
\label{tab:dataset_stats}
\resizebox{\columnwidth}{!}{
\begin{tabular}{@{}lcl@{}}
\toprule
\textbf{Dataset} & \textbf{Sample Count} & \textbf{Tampering Type} \\ \midrule
\rowcolor{gray!10} \multicolumn{3}{l}{\textit{Training Sets}} \\
DocTamper-TrainingSet \cite{qu2023towards} & 120,000 & Com, Spl, Rep \\
FSTS-T \cite{yu2025toward} & 50,000 & Com, Spl, Rem, Ins, Rep \\ \midrule
\rowcolor{gray!10} \multicolumn{3}{l}{\textit{Testing Sets}} \\
DocTamper-TestingSet \cite{qu2023towards} & 30,000 & Com, Spl, Rep \\
DocTamper-FCD \cite{qu2023towards} & 2,000 & Com, Spl, Rep \\
DocTamper-SCD \cite{qu2023towards} & 18,000 & Com, Spl, Rep \\
FSTS-S \cite{yu2025toward} & 5,705 & Com, Spl, Rem, Ins, Rep \\
FSTS-1.5K \cite{yu2025toward} & 1,488 & Com, Spl, Rem, Ins, Rep \\
SACP \cite{comp2020} & 2,005 & Com, Spl, Rem, Rep \\
RIFLC \cite{comp2022} & 4,000 & Com, Spl, Rem \\ \bottomrule
\end{tabular}
}
\end{table}

\subsubsection{\texorpdfstring{Setting of the Expert Model}{Setting of the Expert Model}}
We construct the forensic expert following DITL$^2$ \cite{li2025ditl2}, using DINOv2-ViT-L \cite{oquab2023dinov2} as the backbone and Mask2Former as the segmentation head for pixel-level tampering localization. The expert is supervised with a combination of Binary Cross-Entropy (BCE) and Dice losses, and is trained on the same forensic training set as the corresponding MLLM, i.e., DocTamper-TrainingSet or FSTS-T, without introducing any additional training data. Once trained, only the DINOv2 backbone is retained as the frozen knowledge source in Stage 2: the Mask2Former head is discarded, and no decoder is involved in the internalization process, where the FGRA loss aligns the LLM's shallow representations with the backbone's feature embeddings.

\subsubsection{\texorpdfstring{Setting of the EKI}{Setting of the EKI}}
We adopt Qwen2.5-VL-7B \cite{bai2025qwen2} as our base MLLM. To ensure training efficiency, we employ LoRA \cite{hu2022lora} for fine-tuning, with the rank set to $r=32$ and $\alpha=64$. During inference, the model directly takes the document image and the plain detection instruction (i.e., ``\textit{Detect and localize the tampered text in this image}'') as input, and generates the authenticity judgment together with the bounding boxes of tampered regions via greedy decoding. Notably, neither OCR pre-processing nor the external expert model is involved at inference time.

\subsubsection{\texorpdfstring{Optimization Details}{Optimization Details}}
Both the expert model and the MLLM are trained for 10 epochs using the AdamW \cite{loshchilov2017decoupled} optimizer ($\beta_1=0.9$, $\beta_2=0.999$) with a learning rate of $1e^{-4}$ and a batch size of 8. For the comparison methods, to ensure a fair evaluation, we re-train all models leveraging the unified frameworks of ForensicHub \cite{du2025forensichub} and IMDL-BenCo \cite{ma2024imdl} under the same optimization settings, except for a batch size of 32. Note that all input images are resized to $512 \times 512$ pixels for all methods during both training and testing.

\subsubsection{Evaluation Metrics}
Following established evaluation protocols \cite{qu2023towards, li2025ditl2}, we adopt the Intersection over Union (IoU) and the F1-score as our quantitative metrics. In line with our task reformulation (Sec.~\ref{subsec:reformulation}), our model outputs the bounding boxes of tampered regions, and both metrics are computed at the pixel level within each image, measuring the spatial overlap between the regions covered by the predicted and ground-truth boxes. The per-image scores are then averaged over all images in the test set to obtain the final reported results. Both metrics range from $0$ to $1$, where higher values indicate more accurate localization of the tampered regions. For a fair comparison, the compared methods that output segmentation masks are evaluated in the same manner: each connected component of the predicted mask is converted into its minimum enclosing rectangle, and the resulting boxes are compared against the ground-truth boxes. All compared methods are evaluated under the same protocol.

\subsection{Comparisons with SOTA Methods}
\subsubsection{Competitors}
We evaluate our method against a comprehensive set of state-of-the-art competitors, including general image manipulation localization methods \cite{ma2023iml, guillaro2023trufor, su2025can, zhu2025mesoscopic}, expert TTD models \cite{dong2024robust, song2025cross, luo2025toward, qu2023towards, chen2024enhancing, qu2025revisiting, wong2025adcd, li2025ditl2}, and the recent MLLM-based approach TVSIP \cite{xu2025pixels}. 
All comparison models were re-trained on DocTamper-TrainingSet and FSTS-T under identical settings.

\subsubsection{Quantitative Analysis}
As shown in Tables \ref{tab:results_doctamper} and \ref{tab:results_fsts}, our method consistently achieves the best average performance across all test sets, surpassing competitors by significant margins in both IoU and F1 metrics. 
On the in-domain benchmarks, our method achieves strong localization performance, attaining the best IoU on both Doc-Test and FSTS-S while remaining highly competitive in F1. More importantly, our approach exhibits superior robustness on cross-domain datasets (SACP, RIFLC), whereas traditional methods suffer severe degradation. 
Although the absolute performance on SACP and RIFLC remains relatively low due to the substantial domain gap between the script-generated forgeries in the training data and the manually crafted forgeries in these two benchmarks, our method consistently ranks at or near the top across these settings, demonstrating stronger robustness to substantially different forgery distributions.
This indicates that our model successfully internalizes intrinsic forgery traces rather than overfitting to specific artifacts.
In contrast, general image methods perform poorly due to the lack of text priors. Regarding MLLM-based approaches, the competitor TVSIP primarily focuses on leveraging models to provide interpretability for forensic results. Consequently, both TVSIP and the vanilla Qwen2.5-VL baseline struggle with the Double Mismatch, confirming the necessity of our training strategy.

\begin{table*}[htbp]
\caption{Experimental results when trained on DocTamper-TrainingSet. Doc-Test, FCD, and SCD denote DocTamper-TestingSet, DocTamper-FCD, and DocTamper-SCD, respectively. \textbf{Bold} indicates the best result, and \underline{underline} indicates the second best.}
\label{tab:results_doctamper}
\begin{center}
\resizebox{\textwidth}{!}{
\begin{tabular}{l cccccccccccccccc}
\toprule
\multirow{2}{*}{Method} & \multicolumn{2}{c}{Doc-Test} & \multicolumn{2}{c}{FCD} & \multicolumn{2}{c}{SCD} & \multicolumn{2}{c}{FSTS-S} & \multicolumn{2}{c}{FSTS-1.5K} & \multicolumn{2}{c}{SACP} & \multicolumn{2}{c}{RIFLC} & \multicolumn{2}{c}{Average} \\
\cmidrule(lr){2-3} \cmidrule(lr){4-5} \cmidrule(lr){6-7} \cmidrule(lr){8-9} \cmidrule(lr){10-11} \cmidrule(lr){12-13} \cmidrule(lr){14-15} \cmidrule(lr){16-17}
 & IoU & F1 & IoU & F1 & IoU & F1 & IoU & F1 & IoU & F1 & IoU & F1 & IoU & F1 & IoU & F1 \\
\midrule
IML-ViT \cite{ma2023iml} & 0.372 & 0.430 & 0.421 & 0.502 & 0.383 & 0.455 & 0.159 & 0.232 & 0.238 & 0.295 & 0.057 & 0.090 & 0.052 & 0.079 & 0.240 & 0.297 \\
TruFor \cite{guillaro2023trufor} & 0.173 & 0.208 & 0.164 & 0.239 & 0.194 & 0.249 & 0.030 & 0.047 & 0.142 & 0.190 & 0.025 & 0.039 & 0.025 & 0.039 & 0.108 & 0.144 \\
SparseViT \cite{su2025can} & 0.636 & 0.714 & 0.437 & 0.500 & 0.495 & 0.601 & 0.082 & 0.114 & 0.378 & 0.459 & 0.028 & 0.044 & 0.033 & 0.047 & 0.298 & 0.354 \\
Mesorch \cite{zhu2025mesoscopic} & 0.463 & 0.526 & 0.462 & 0.523 & 0.356 & 0.433 & 0.167 & 0.221 & 0.178 & 0.223 & 0.032 & 0.046 & 0.018 & 0.030 & 0.239 & 0.286 \\
\midrule
TIFDM \cite{dong2024robust} & 0.496 & 0.556 & 0.368 & 0.427 & 0.387 & 0.467 & 0.113 & 0.167 & 0.074 & 0.104 & 0.018 & 0.032 & 0.025 & 0.039 & 0.212 & 0.256 \\
CAFTB \cite{song2025cross} & 0.685 & 0.746 & 0.542 & 0.599 & 0.544 & 0.643 & 0.286 & 0.364 & 0.241 & 0.294 & 0.056 & 0.085 & 0.077 & 0.110 & 0.347 & 0.406 \\
ASCFormer \cite{luo2025toward} & 0.810 & 0.744 & 0.670 & 0.750 & 0.608 & 0.670 & 0.280 & 0.311 & 0.172 & 0.266 & 0.111 & 0.112 & 0.057 & 0.099 & 0.387 & 0.422 \\
DTD \cite{qu2023towards} & 0.826 & 0.875 & 0.784 & 0.833 & 0.749 & 0.822 & 0.066 & 0.097 & 0.346 & 0.405 & 0.021 & 0.034 & 0.035 & 0.054 & 0.404 & 0.446 \\
FFDN \cite{chen2024enhancing} & \underline{0.877} & \textbf{0.917} & \underline{0.833} & \underline{0.880} & \underline{0.789} & \underline{0.858} & 0.141 & 0.193 & \underline{0.393} & \underline{0.465} & 0.028 & 0.045 & 0.040 & 0.062 & 0.443 & 0.488 \\
DAF \cite{qu2025revisiting} & 0.657 & 0.720 & 0.492 & 0.543 & 0.235 & 0.296 & 0.128 & 0.185 & 0.193 & 0.246 & 0.042 & 0.057 & 0.034 & 0.054 & 0.254 & 0.300 \\
ADCD-Net \cite{wong2025adcd} & 0.835 & 0.814 & 0.739 & 0.821 & 0.641 & 0.733 & 0.282 & 0.353 & 0.268 & 0.426 & \underline{0.125} & \underline{0.135} & \underline{0.089} & \underline{0.111} & 0.426 & 0.485 \\
DITL$^2$ \cite{li2025ditl2} & 0.839 & 0.883 & 0.817 & 0.876 & 0.680 & 0.774 & \textbf{0.368} & \textbf{0.474} & 0.361 & 0.452 & 0.034 & 0.060 & 0.049 & 0.079 & \underline{0.450} & \underline{0.514} \\
\midrule
Qwen2.5-VL \cite{bai2025qwen2} & 0.696 & 0.702 & 0.313 & 0.616 & 0.411 & 0.606 & 0.146 & 0.313 & 0.106 & 0.216 & 0.018 & 0.021 & 0.019 & 0.030 & 0.244 & 0.358 \\
TVSIP \cite{xu2025pixels} & 0.726 & 0.768 & 0.258 & 0.583 & 0.542 & 0.669 & 0.089 & 0.164 & 0.069 & 0.193 & 0.056 & 0.106 & 0.053 & 0.101 & 0.256 & 0.369 \\
Ours & \textbf{0.890} & \underline{0.900} & \textbf{0.875} & \textbf{0.921} & \textbf{0.822} & \textbf{0.893} & \underline{0.317} & \underline{0.392} & \textbf{0.400} & \textbf{0.584} & \textbf{0.131} & \textbf{0.151} & \textbf{0.101} & \textbf{0.123} & \textbf{0.505} & \textbf{0.566} \\
\bottomrule
\end{tabular}
}
\end{center}
\end{table*}

\begin{table*}[htbp]
\caption{Experimental results when trained on \textbf{FSTS-T}. Doc-Test, FCD, and SCD denote DocTamper-TestingSet, DocTamper-FCD, and DocTamper-SCD, respectively. \textbf{Bold} indicates the best result, and \underline{underline} indicates the second best.}
\label{tab:results_fsts}
\begin{center}
\resizebox{\textwidth}{!}{
\begin{tabular}{l cccccccccccccccc}
\toprule
\multirow{2}{*}{Method} & \multicolumn{2}{c}{FSTS-S} & \multicolumn{2}{c}{FSTS-1.5K} & \multicolumn{2}{c}{Doc-Test} & \multicolumn{2}{c}{FCD} & \multicolumn{2}{c}{SCD} & \multicolumn{2}{c}{SACP} & \multicolumn{2}{c}{RIFLC} & \multicolumn{2}{c}{Average} \\
\cmidrule(lr){2-3} \cmidrule(lr){4-5} \cmidrule(lr){6-7} \cmidrule(lr){8-9} \cmidrule(lr){10-11} \cmidrule(lr){12-13} \cmidrule(lr){14-15} \cmidrule(lr){16-17}
 & IoU & F1 & IoU & F1 & IoU & F1 & IoU & F1 & IoU & F1 & IoU & F1 & IoU & F1 & IoU & F1 \\
\midrule
IML-ViT \cite{ma2023iml} & 0.358 & 0.464 & 0.069 & 0.107 & 0.071 & 0.106 & 0.114 & 0.169 & 0.111 & 0.160 & 0.056 & 0.096 & 0.060 & 0.098 & 0.120 & 0.171 \\
TruFor \cite{guillaro2023trufor} & 0.341 & 0.431 & 0.199 & 0.270 & 0.109 & 0.142 & 0.088 & 0.129 & 0.086 & 0.118 & 0.100 & 0.157 & 0.086 & 0.131 & 0.144 & 0.197 \\
SparseViT \cite{su2025can} & 0.211 & 0.285 & 0.454 & 0.526 & 0.139 & 0.173 & 0.042 & 0.057 & 0.118 & 0.160 & 0.048 & 0.077 & 0.060 & 0.090 & 0.153 & 0.195 \\
Mesorch \cite{zhu2025mesoscopic} & 0.572 & 0.675 & 0.443 & 0.529 & 0.211 & 0.248 & 0.202 & 0.263 & 0.185 & 0.230 & 0.084 & 0.131 & 0.102 & 0.146 & 0.257 & 0.317 \\
\midrule
TIFDM \cite{dong2024robust} & 0.486 & 0.592 & 0.140 & 0.196 & 0.105 & 0.142 & 0.057 & 0.090 & 0.094 & 0.133 & 0.048 & 0.080 & 0.057 & 0.090 & 0.141 & 0.189 \\
CAFTB \cite{song2025cross} & 0.687 & 0.781 & 0.391 & 0.459 & 0.266 & 0.311 & 0.350 & 0.423 & 0.275 & 0.337 & 0.091 & 0.137 & \textbf{0.109} & \textbf{0.153} & 0.310 & 0.371 \\
ASCFormer \cite{luo2025toward} & 0.690 & 0.761 & 0.513 & 0.547 & 0.451 & 0.375 & 0.492 & 0.594 & 0.309 & 0.400 & 0.108 & 0.133 & 0.062 & 0.080 & 0.375 & 0.413 \\
DTD \cite{qu2023towards} & 0.227 & 0.284 & 0.566 & 0.634 & \underline{0.519} & \textbf{0.589} & 0.571 & 0.642 & \textbf{0.471} & \underline{0.567} & 0.055 & 0.086 & 0.067 & 0.097 & 0.354 & 0.414 \\
FFDN \cite{chen2024enhancing} & 0.142 & 0.186 & 0.537 & 0.609 & 0.478 & \underline{0.550} & 0.532 & 0.612 & 0.417 & 0.512 & 0.069 & 0.108 & 0.072 & 0.107 & 0.321 & 0.383 \\
DAF \cite{qu2025revisiting} & 0.440 & 0.531 & 0.515 & 0.567 & 0.471 & 0.517 & 0.542 & 0.627 & 0.128 & 0.227 & \underline{0.124} & \underline{0.139} & 0.101 & 0.136 & 0.331 & 0.392 \\
ADCD-Net \cite{wong2025adcd} & \underline{0.742} & \underline{0.812} & 0.426 & 0.528 & 0.373 & 0.342 & \underline{0.598} & \underline{0.671} & 0.328 & 0.446 & 0.083 & 0.091 & 0.103 & 0.127 & 0.379 & 0.431 \\
DITL$^2$ \cite{li2025ditl2} & 0.662 & 0.745 & \textbf{0.653} & \textbf{0.790} & 0.469 & 0.461 & 0.483 & 0.592 & 0.361 & 0.457 & 0.122 & 0.128 & 0.105 & 0.115 & \underline{0.408} & \underline{0.470} \\
\midrule
Qwen2.5-VL \cite{bai2025qwen2} & 0.658 & 0.760 & 0.199 & 0.368 & 0.215 & 0.292 & 0.283 & 0.441 & 0.141 & 0.353 & 0.029 & 0.060 & 0.038 & 0.063 & 0.223 & 0.334 \\
TVSIP \cite{xu2025pixels} & 0.687 & 0.753 & 0.202 & 0.402 & 0.318 & 0.389 & 0.224 & 0.320 & 0.117 & 0.301 & 0.072 & 0.124 & 0.069 & 0.114 & 0.241 & 0.343 \\
Ours & \textbf{0.753} & \textbf{0.838} & \underline{0.648} & \underline{0.741} & \textbf{0.529} & {0.541} & \textbf{0.612} & \textbf{0.776} & \underline{0.435} & \textbf{0.633} & \textbf{0.128} & \textbf{0.147} & \underline{0.108} & \underline{0.150} & \textbf{0.459} & \textbf{0.547} \\
\bottomrule
\end{tabular}
}
\end{center}
\end{table*}

\subsubsection{Qualitative Analysis}
\begin{figure*}[t]
\centering
\includegraphics[width=\textwidth]{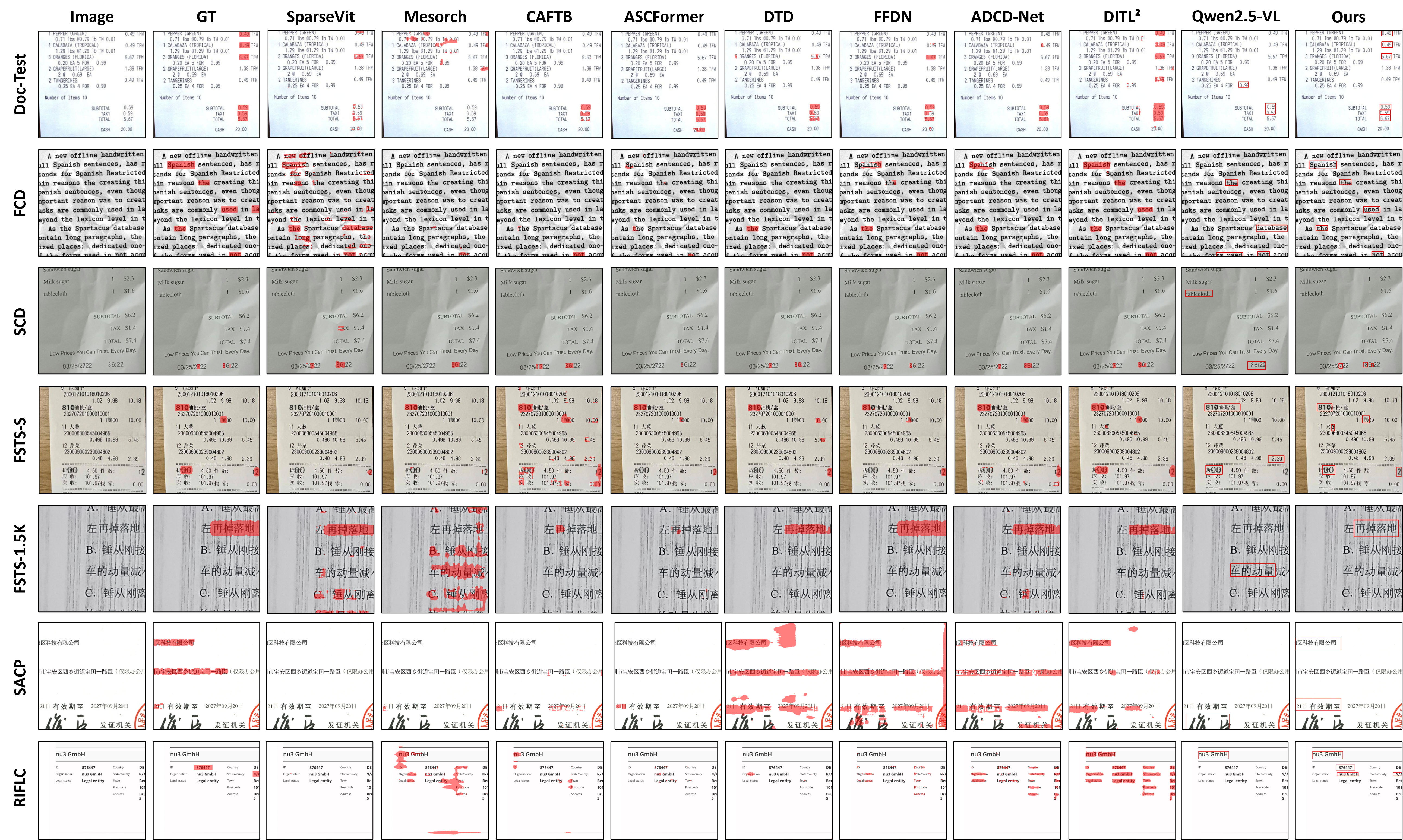}
\caption{Qualitative comparison of different methods trained on DocTamper-TrainingSet. Compared to the blind spots of general image methods, the fragmented masks/false positives of expert TTD models, and the coarse bounding boxes of the standard MLLM baseline, our EKI method generates highly precise and compact localization results.}
\label{fig:qualitative_results}
\end{figure*}

To further demonstrate the superiority of our proposed method, we present a qualitative comparison of models trained on DocTamper-TrainingSet, as illustrated in Fig. \ref{fig:qualitative_results}. 
As can be clearly observed, general image manipulation localization methods often exhibit large blind spots due to the lack of document-specific priors. Expert TTD models, when confronted with complex backgrounds, frequently output blurry and fragmented prediction masks or trigger severe regional false positives. Meanwhile, the standard MLLM baseline frequently suffers from semantic hallucinations, resulting in misaligned or coarse bounding boxes. In stark contrast, our proposed EKI algorithm consistently generates extremely compact and precise bounding boxes, fully demonstrating its superior spatial grounding and artifact perception capabilities.

\subsubsection{Inference Efficiency Analysis}
In Table \ref{tab:inference_time}, we report the inference time per image for our method and various state-of-the-art competitors. 
As expected, expert models (e.g., CAFTB, DTD, FFDN) exhibit lower latency due to their significantly smaller parameter counts (typically in the range of millions). In contrast, MLLM-based approaches inherently require more processing time due to their massive architectures. However, compared to the recent MLLM competitor TVSIP, our method is notably more efficient. 
While the inference time is naturally higher than that of compact experts, it remains within a practical range for high-precision document forensic applications where accuracy and visual reasoning are prioritized over real-time processing. 
Crucially, our EKI paradigm directly internalizes the expert model's capabilities into the MLLM's own parameters. Because the external expert model is completely discarded after the training stage, our method operates with an inference speed nearly identical to that of the vanilla Qwen2.5-VL baseline, introducing absolutely no additional computational overhead during inference. The increased latency compared to expert models is a necessary and acceptable trade-off for the significantly superior detection precision and cross-domain generalization capabilities demonstrated in our experiments.

\begin{table}[htbp]
\centering
\caption{Comparison of inference time per image. During the evaluation, each input image is resized to a fixed resolution of $512 \times 512$.}
\label{tab:inference_time}
\resizebox{\columnwidth}{!}{
\begin{tabular}{@{}llc@{}}
\toprule
\textbf{Method} & \textbf{Publication / Year} & \textbf{Inference Time (ms)} \\ \midrule
IML-ViT \cite{ma2023iml} & arXiv 2023 & 247.3 \\
TruFor \cite{guillaro2023trufor} & CVPR 2023 & 314.9 \\
SparseViT \cite{su2025can} & AAAI 2025 & 122.7 \\
Mesorch \cite{zhu2025mesoscopic} & AAAI 2025 & 55.7 \\
TIFDM \cite{dong2024robust} & IEEE TCE 2024 & 282.1 \\
CAFTB \cite{song2025cross} & ACM TOMM 2025 & 36.1 \\
ASCFormer \cite{luo2025toward} & PR 2025 & 146.0 \\
DTD \cite{qu2023towards} & CVPR 2023 & 52.4 \\
FFDN \cite{chen2024enhancing} & ECCV 2024 & 77.6 \\
DAF \cite{qu2025revisiting} & AAAI 2025 & 248.5 \\
ADCD-Net \cite{wong2025adcd} & ICCV 2025 & 135.2 \\
DITL$^2$ \cite{li2025ditl2} & ACM MM 2025 & 350.4 \\ \midrule
Qwen2.5-VL \cite{bai2025qwen2} & arXiv 2025 & 823.8 \\
TVSIP \cite{xu2025pixels} & ACM MM 2025 & 1063.6 \\
Ours & - & 828.6 \\ \bottomrule
\end{tabular}
}
\end{table}

\subsection{Ablation Studies}
We conduct ablation studies on DocTamper-TrainingSet to investigate the contribution of each component.

\subsubsection{Component Analysis}
As reported in Table \ref{tab:ablation_study}, the vanilla baseline yields suboptimal results, confirming the inherent limitations of general MLLMs in fine-grained forensics.
Strategies designed for Spatial Precision Enhancement—Text-Focused and Image-Focused—individually lead to noticeable improvements, with their combination further enhancing localization quality.
However, the most critical driver is the FGRA loss. Even without Stage 1 strategies, adding FGRA results in a dramatic performance leap, confirming that Forensic Artifact Perception is paramount for distinguishing genuine from forged content. In this FGRA-only setting, the model is trained in a single stage: the vision encoder is kept frozen at its pre-trained weights, and the LLM and projector are fine-tuned with $\mathcal{L}_{LM} + \mathcal{L}_{FGRA}$, which is equivalent to applying Stage 2 directly without the Stage 1 spatial-focus training.
Notably, the FGRA-only variant already surpasses the combination of both Stage 1 strategies, indicating that the Perceptual Granularity Mismatch is the dominant bottleneck, while precise spatial focus acts as an amplifier of the internalized perception.
The full model achieves the best results, demonstrating the mutually reinforcing effect: precise spatial focus lays the spatial foundation, while internalized forensic knowledge enables accurate discrimination.

\begin{table*}[htbp]
\caption{Ablation study of different components on DocTamper-TrainingSet. ``TF'', ``IF'', and ``FGRA'' represent the Text-Focused, Image-Focused, and Forensic-General Representation Alignment, respectively. Doc-Test, FCD, and SCD denote DocTamper-TestingSet, DocTamper-FCD, and DocTamper-SCD. \textbf{Bold} indicates the best result, and \underline{underline} indicates the second best.}
\label{tab:ablation_study}
\begin{center}
\resizebox{\textwidth}{!}{
\begin{tabular}{ccc cccccccccccccccc}
\toprule
\multicolumn{3}{c}{Method} & \multicolumn{2}{c}{Doc-Test} & \multicolumn{2}{c}{FCD} & \multicolumn{2}{c}{SCD} & \multicolumn{2}{c}{FSTS-S} & \multicolumn{2}{c}{FSTS-1.5K} & \multicolumn{2}{c}{SACP} & \multicolumn{2}{c}{RIFLC} & \multicolumn{2}{c}{Average} \\
\cmidrule(lr){1-3} \cmidrule(lr){4-5} \cmidrule(lr){6-7} \cmidrule(lr){8-9} \cmidrule(lr){10-11} \cmidrule(lr){12-13} \cmidrule(lr){14-15} \cmidrule(lr){16-17} \cmidrule(lr){18-19}
TF & IF & FGRA & IoU & F1 & IoU & F1 & IoU & F1 & IoU & F1 & IoU & F1 & IoU & F1 & IoU & F1 & IoU & F1 \\
\midrule
$\times$ & $\times$ & $\times$ & 0.696 & 0.702 & 0.313 & 0.616 & 0.411 & 0.606 & 0.146 & 0.313 & 0.106 & 0.216 & 0.018 & 0.021 & 0.019 & 0.030 & 0.244 & 0.358 \\
\checkmark & $\times$ & $\times$ & 0.776 & 0.783 & 0.367 & 0.648 & 0.485 & 0.659 & 0.150 & 0.303 & 0.148 & 0.266 & 0.021 & 0.029 & 0.022 & 0.039 & 0.281 & 0.389 \\
$\times$ & \checkmark & $\times$ & 0.837 & 0.822 & 0.471 & 0.673 & 0.616 & 0.772 & 0.161 & 0.316 & 0.150 & 0.275 & 0.041 & 0.075 & 0.034 & 0.062 & 0.330 & 0.428 \\
\checkmark & \checkmark & $\times$ & \underline{0.870} & \underline{0.860} & 0.517 & 0.710 & 0.635 & 0.781 & 0.180 & 0.326 & 0.195 & 0.325 & 0.058 & 0.086 & 0.043 & 0.071 & 0.357 & 0.451 \\
$\times$ & $\times$ & \checkmark & 0.827 & 0.829 & \underline{0.861} & \underline{0.914} & \underline{0.722} & \underline{0.843} & \underline{0.242} & \underline{0.329} & \underline{0.351} & \underline{0.534} & \underline{0.093} & \underline{0.144} & \underline{0.090} & \underline{0.118} & \underline{0.455} & \underline{0.530} \\
\checkmark & \checkmark & \checkmark & \textbf{0.890} & \textbf{0.900} & \textbf{0.875} & \textbf{0.921} & \textbf{0.822} & \textbf{0.893} & \textbf{0.317} & \textbf{0.392} & \textbf{0.400} & \textbf{0.584} & \textbf{0.131} & \textbf{0.151} & \textbf{0.101} & \textbf{0.123} & \textbf{0.505} & \textbf{0.566} \\
\bottomrule
\end{tabular}
}
\end{center}
\end{table*}

\subsubsection{Analysis of the Alignment Layer Depth}
\begin{figure}[t]
\centering
\includegraphics[width=\columnwidth]{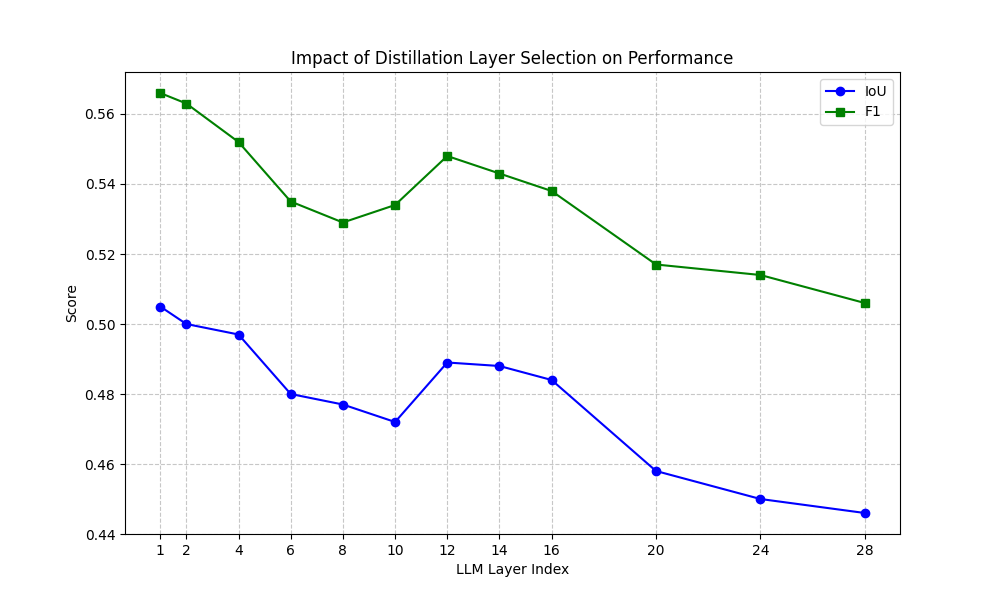}
\caption{Ablation study on the selection of LLM layers in FGRA. Aligning with the shallowest layer ($l=1$) yields optimal performance, whereas deeper layers show a clear declining trend.}
\label{fig:layer_ablation}
\end{figure}

\begin{figure}[t]
\centering
\includegraphics[width=\columnwidth]{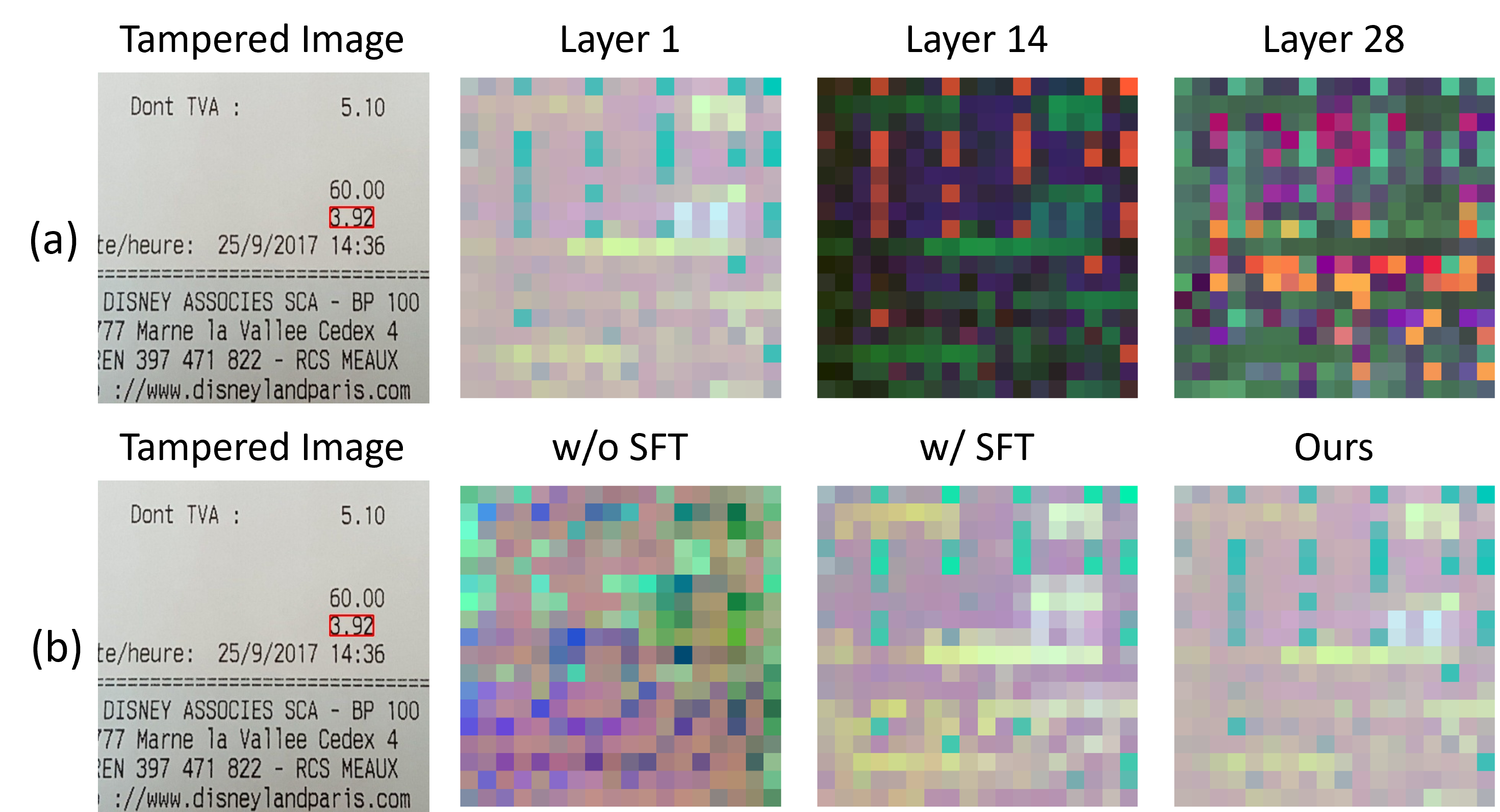}
\caption{Visualization results. (a) Visualizations of outputs from different layers of the LLM; (b) Visualizations of the first layer outputs from the original Qwen2.5-VL-7B, after supervised fine-tuning, and our proposed method.}
\label{fig:visual_ablation}
\end{figure}

To validate the design of the FGRA loss, we explore the impact of alignment layer $l$ selection. 
As illustrated in Fig. \ref{fig:layer_ablation}, we observe a clear declining trend in detection performance as the alignment layer moves deeper. Specifically, aligning with the first layer yields the optimal performance, while deeper layers lead to significant degradation.
This phenomenon can be attributed to the hierarchical nature of LLMs: shallow layers act as the initial visual-linguistic interface, retaining more structural and textural details essential for forensics; in contrast, deep layers evolve into highly abstract, semantic-dominant representations where subtle tampering traces are diluted.
To qualitatively verify this, we visualize the layer-wise visual-token representations following the PCA-based feature visualization protocol of DINOv2 \cite{oquab2023dinov2}, which has also been adopted in recent MLLM representation alignment studies \cite{yoon2025visual}. Specifically, for each LLM layer, we project the hidden states of the visual tokens onto their top three principal components, which are min-max normalized and rendered as RGB channels over the visual-token grid. Unlike DINOv2, PCA is applied to all visual tokens without foreground separation, since tampering evidence in document images may reside in any region rather than a salient foreground. Tokens with similar representations thus share similar colors, and a color boundary indicates that the layer separates the corresponding regions in its feature space.
This finding is further corroborated by the visualizations in Fig. \ref{fig:visual_ablation}. Fig. \ref{fig:visual_ablation}(a) shows that forensic signals are distinct in shallow feature maps but fade in deeper layers, justifying our choice of $l=1$. Furthermore, Fig. \ref{fig:visual_ablation}(b) demonstrates the efficacy of this alignment: while the original Qwen2.5-VL and standard Supervised Fine-Tuning (SFT) models show noisy or blind activation patterns, our method exhibits distinctive feature responses precisely on the manipulated regions, confirming the successful internalization of the expert's discriminative capability. 

\subsubsection{Analysis of Different Fusion Strategies}
To investigate how forensic expertise can be most effectively incorporated into MLLMs, we compare four representative paradigms: (i) \textit{Baseline}, a standard fine-tuned MLLM without forensic priors; (ii) \textit{Expert Model}, a DINOv2-ViT-L forensic detector trained following DITL$^2$ \cite{li2025ditl2}; (iii) \textit{ER} (External Reliance), which retains the expert during inference and combines its prediction with the MLLM through late fusion; and (iv) our \textit{EKI} (Expert Knowledge Internalization), which transfers the expert’s forensic perception into the MLLM and removes the expert at inference. For a fair comparison, the same fine-tuned DINOv2-ViT-L expert is used in paradigms (ii)–(iv).
As shown in Table~\ref{tab:fusion_strategy}, the Baseline performs the worst, highlighting the limited sensitivity of a generic MLLM to subtle forensic artifacts. The standalone Expert Model provides a clear improvement, confirming the benefit of specialized forensic representations. However, ER still fails to outperform the expert itself, suggesting that simply exposing the MLLM to external expert predictions does not effectively integrate the complementary strengths of the two models. In contrast, EKI achieves the best overall performance, reaching an average IoU/F1 of $0.505/0.566$, while requiring no external expert at inference. These results indicate that transferring forensic knowledge into the MLLM’s own representations is more effective than relying on an externally coupled expert, and further supports the motivation of our internalization paradigm.

\begin{table}[t]
\centering
\caption{Comparison of different paradigms for integrating forensic expertise into the MLLM. All metrics are averaged over the test sets. ``ER'' and ``EKI'' denote External Reliance and our Expert Knowledge Internalization, respectively. \textbf{Bold} indicates the best result, and \underline{underline} the second best.}
\label{tab:fusion_strategy}
\begin{tabular}{lcc}
\toprule
Paradigm & IoU & F1 \\
\midrule
Baseline (fine-tuned MLLM) & 0.244 & 0.358 \\
Expert Model & \underline{0.450} & \underline{0.514} \\
ER (External Reliance) & 0.428 & 0.501 \\
EKI (Ours) & \textbf{0.505} & \textbf{0.566} \\
\bottomrule
\end{tabular}
\end{table}

\subsubsection{Analysis of the Base MLLM}
To verify the versatility of our proposed method, we conduct an ablation study across different base MLLM architectures. We compare the performance of standard fine-tuning versus our EKI framework using three different models: LLaVA-1.5-7B\cite{liu2024improved}, Qwen2.5-VL 7B\cite{bai2025qwen2}, and the more recent Qwen3-VL 8B\cite{bai2025qwen3}. All models are trained on DocTamper-TrainingSet.
As shown in Table \ref{tab:base_model_ablation}, the Double Mismatch is a universal bottleneck for general MLLMs. Regardless of the base model's inherent capacity, standard fine-tuning invariably yields suboptimal forensic localization, particularly exhibiting poor generalization in cross-domain scenarios. While advancing the base architecture from LLaVA-1.5 to Qwen3-VL naturally yields a slight baseline improvement due to enhanced general vision-language capabilities, the inherent blindness to fine-grained forensic artifacts remains largely unresolved.
Crucially, the application of our EKI framework brings substantial and consistent performance leaps across all tested architectures. Furthermore, adopting a stronger base model under our paradigm establishes a new state-of-the-art performance ceiling, confirming that our internalization process can synergize with the model's intrinsic reasoning power. These consistent improvements clearly demonstrate that our strategy is model-agnostic. By systematically internalizing external expert knowledge, EKI effectively bridges the granularity mismatch, serving as a plug-and-play forensic capability booster for broad MLLM families.

\begin{table*}[htbp]
\caption{Ablation study on the choice of different base MLLMs. All models are trained on DocTamper-TrainingSet. ``Baseline'' refers to standard fine-tuning, and ``Ours'' indicates the application of the proposed EKI framework. \textbf{Bold} indicates the best result, and \underline{underline} indicates the second best.}
\label{tab:base_model_ablation}
\begin{center}
\resizebox{\textwidth}{!}{
\begin{tabular}{ll cccccccccccccccc}
\toprule
\multirow{2}{*}{Base MLLM} & \multirow{2}{*}{Method} & \multicolumn{2}{c}{Doc-Test} & \multicolumn{2}{c}{FCD} & \multicolumn{2}{c}{SCD} & \multicolumn{2}{c}{FSTS-S} & \multicolumn{2}{c}{FSTS-1.5K} & \multicolumn{2}{c}{SACP} & \multicolumn{2}{c}{RIFLC} & \multicolumn{2}{c}{Average} \\
\cmidrule(lr){3-4} \cmidrule(lr){5-6} \cmidrule(lr){7-8} \cmidrule(lr){9-10} \cmidrule(lr){11-12} \cmidrule(lr){13-14} \cmidrule(lr){15-16} \cmidrule(lr){17-18}
 & & IoU & F1 & IoU & F1 & IoU & F1 & IoU & F1 & IoU & F1 & IoU & F1 & IoU & F1 & IoU & F1 \\
\midrule
\multirow{2}{*}{LLaVA-1.5-7B} 
& Baseline & 0.648 & 0.674 & 0.264 & 0.517 & 0.360 & 0.513 & 0.074 & 0.198 & 0.207 & 0.347 & 0.011 & 0.020 & 0.011 & 0.031 & 0.225 & 0.328 \\
& Ours & 0.817 & 0.861 & 0.788 & 0.878 & \underline{0.814} & \underline{0.891} & \underline{0.319} & \underline{0.399} & 0.363 & 0.540 & 0.122 & 0.140 & \underline{0.115} & \underline{0.128} & 0.477 & 0.548 \\
\midrule
\multirow{2}{*}{Qwen2.5-VL 7B} 
& Baseline & 0.696 & 0.702 & 0.313 & 0.616 & 0.411 & 0.606 & 0.146 & 0.313 & 0.106 & 0.216 & 0.018 & 0.021 & 0.019 & 0.030 & 0.244 & 0.358 \\
& Ours & \textbf{0.890} & \textbf{0.900} & \textbf{0.875} & \textbf{0.921} & \textbf{0.822} & \textbf{0.893} & 0.317 & 0.392 & \underline{0.400} & \underline{0.584} & \underline{0.131} & \underline{0.151} & 0.101 & 0.123 & \underline{0.505} & \underline{0.566} \\
\midrule
\multirow{2}{*}{Qwen3-VL 8B} 
& Baseline & 0.724 & 0.730 & 0.323 & 0.627 & 0.425 & 0.617 & 0.150 & 0.316 & 0.118 & 0.226 & 0.019 & 0.039 & 0.019 & 0.032 & 0.254 & 0.370 \\
& Ours & \underline{0.870} & \underline{0.888} & \underline{0.871} & \underline{0.909} & 0.806 & 0.865 & \textbf{0.369} & \textbf{0.479} & \textbf{0.463} & \textbf{0.600} & \textbf{0.152} & \textbf{0.160} & \textbf{0.135} & \textbf{0.138} & \textbf{0.524} & \textbf{0.577} \\
\bottomrule
\end{tabular}
}
\end{center}
\end{table*}

\section{Limitations}
Since existing MLLM-based TTD methods still fall short in detection performance, our work first focuses on improving the model's ability to detect and localize tampered text. Consequently, it does not advance the human-readable explanation of tampering evidence emphasized by prior MLLM-based methods \cite{qu2024textsleuth, xu2025pixels}. We believe this capability is promising for improving the reliability and trustworthiness of forensic systems, and leave its integration into our framework for future work.
Moreover, our bounding-box formulation is sufficient for the primary objective considered in this work, i.e., identifying which text region has been manipulated, while being naturally compatible with the autoregressive output paradigm of MLLMs. Nevertheless, bounding boxes do not delineate the exact shape of irregular tampered regions, and our current formulation is therefore not intended for applications that explicitly require pixel-accurate boundary localization.

\section{Conclusion}
In this paper, we investigate how to equip MLLMs with expert forensic perception for tampered text detection. Our experiments show that simply coupling an MLLM with an external forensic expert does not fully exploit their complementary strengths and can even underperform the standalone expert. We further identify the Double Mismatch that limits direct adaptation of MLLMs to fine-grained forensics: the Spatial Precision Mismatch between coarse visual tokens and tiny tampered regions, and the Perceptual Granularity Mismatch between semantics-oriented representations and subtle forensic artifacts.
To address these challenges, we propose Expert Knowledge Internalization (EKI), which uses the forensic expert as a teacher to guide the MLLM’s own representation learning. EKI follows a progressive two-stage design: Stage 1 improves spatial focus through Text-Focused and Image-Focused training, while Stage 2 transfers fine-grained forensic perception through the proposed FGRA loss. The expert is used only during training and is removed at inference, so no external expert is required during deployment.
Extensive experiments show that EKI achieves state-of-the-art performance across diverse document benchmarks and consistently improves different base MLLMs. Compared with external reliance, EKI provides stronger cross-domain generalization by integrating forensic perception into the MLLM itself. These results suggest that expert-guided representation learning is a promising way to equip general-purpose MLLMs with specialized perceptual capabilities for forensic tasks.

\bibliographystyle{IEEEtran}
\bibliography{IEEEabrv,citation_paper}

\end{document}